\documentclass[sn-basic]{sn-jnl}
  
\usepackage{amsmath,amsfonts,amssymb,amsthm}
\usepackage{circuitikz}
\usepackage{caption}
 \usepackage{subfig}
 \usepackage{multirow}
\usepackage{pifont}
\usepackage{times}
\usepackage{graphicx}
\usepackage{color}
\usepackage{booktabs}
\usepackage{adjustbox}
\usepackage{setspace}
\usepackage{float}
\usepackage{comment}
\usepackage{xcolor}
\usepackage{rotating}
\usepackage{bbm}
\usepackage{latexsym}

\usepackage{tikz}
\usepackage{float}
\usepackage{pgfplots}
\usepackage{pgfplotstable}
\pgfplotsset{
	compat=newest,
}
\usepgfplotslibrary{fillbetween}

\usetikzlibrary{external}
\usetikzlibrary{patterns}
\newcommand{\abl}[2]{\frac{\mathrm{d}#1}{\mathrm{d}#2}}
\definecolor{brinkpink}{rgb}{0.98, 0.38, 0.5}
\definecolor{darkblue}{rgb}{0.0, 0.0, 0.55}
\definecolor{bluemat}{rgb}{0.26, 0.33, 0.55}
\definecolor{redmat}{rgb}{0.6, 0.25, 0.21}
\definecolor{greenmat}{rgb}{0.28, 0.63, 0.3}

\newcommand{\one}{($i$) }
\newcommand{\two}{($ii$) }

\usepackage{graphicx}%
\usepackage{multirow}%
\usepackage{amsmath,amssymb,amsfonts}%
\usepackage{amsthm}%
\usepackage{mathrsfs}%
\usepackage{xcolor}%
\usepackage{textcomp}%
\usepackage{manyfoot}%
\usepackage{booktabs}%
\usepackage{algorithm}%
\usepackage{algorithmic}%
\usepackage{listings}%

\theoremstyle{thmstyleone}%
\theoremstyle{thmstyletwo}%

\theoremstyle{thmstylethree}%

\begin{document}

\title[Article Title]{Article Title}
	\title[]{Forward Direct Feedback Alignment for Online Gradient Estimates of Spiking Neural Networks}


\author[1]{\fnm{Florian} \sur{Bacho}}\email{f.bacho@outlook.fr}

\author*[1]{\fnm{Dominique} \sur{Chu}}\email{d.f.chu@kent.ac.uk}

\affil*[1]{\orgdiv{CEMS, School of Computing}, \orgname{University of Kent}, \orgaddress{\street{Kennedy Building}, \city{Canterbury}, \postcode{CT2 7NF}, \state{Kent}, \country{United Kingdom}}}


\abstract{
There is an interest in  finding energy efficient alternatives to current state of the art neural network training algorithms. Spiking neural network are a promising approach, because they can be simulated energy efficiently on neuromorphic hardware platforms. However, these platforms come with limitations on the design of the training algorithm. Most importantly, backpropagation cannot be implemented on those.  We propose a novel neuromorphic algorithm,  the \textit{Spiking Forward Direct Feedback Alignment} (SFDFA) algorithm, an adaption of \textit{Forward Direct Feedback Alignment} to train  SNNs. SFDFA  estimates the weights between output and hidden neurons as feedback connections. The main contribution of this paper is to describe  how exact local gradients of spikes can be computed in an online manner while taking into account the intra-neuron dependencies between post-synaptic spikes and derive a dynamical system for neuromorphic hardware compatibility. We  compare the SFDFA algorithm with a number of competitor algorithms  and show that the proposed algorithm achieves higher performance and convergence rates.
}

\keywords{Spiking Neural Networks, Tradeoffs, Time-To-First-Spike, Error Backpropagation }



\maketitle

\section{Introduction}

The backpropagation algorithm (BP) \citep{error_backpropagation}  underpins a good part of modern neural network (NN) based AI. BP-based training algorithms continue to be the state of the art in many areas of machine learning ranging from benchmark problems such as the MNIST dataset\cite{mnist} to the most recent transformer-based architectures \citep{dfa_scales_to_modern_dl}.      While its success is undeniable, BP has some disadvantages. The main one is that BP is computationally expensive and  relies on sequential processing of layers during both the forward and backward pass, limiting its scope for parallelisation. This is sometimes called {\em backward locking} \citep{decoupled_parallel_bp}. 
\par
Furthermore, under BP the update of any particular connection weight in a neural network requires global information about the entire network. This entails intense data processing needs \citep{sparse_dfa} and consequently high energy consumption \cite{sparse_dfa,dfa_cnn_online_learning_processor}. Most importantly,  BP  is also not compatible with  neuromorphic  hardware platforms \citep{event_driven_random_bp}, such as Loihi \citep{loihi} or SpiNNaker \citep{spinnaker}, because neuromorphic hardware cannot save the past states of neurons or reverse time. 
\par
In the light of this, there has been some recent interest in alternatives to BP  that alleviate these issues.  In the context of SNNs, random feedback learning, and particularly the  {\em Direct Feedback Alignment} (DFA) algorithm is an interesting alternative to BP for online gradient approximation \citep{erbp_dfa, eprop_dfa, eprop_dfa_spinnaker, st_dfa, emstdp_dfa_2, emstdp_dfa}. For example, the \textit{Event-Driven Random Backpropagation} (eRBP) \citep{erbp_dfa} and the \textit{e-prop} \citep{eprop_dfa, eprop_dfa_spinnaker} algorithms use random feedback connections to send error signals to hidden neurons at every time steps, thus continuously accumulating gradients during the inference. Unlike BP, DFA algorithms can be run efficiently on neuromorphic hardware. However, these algorithms constantly project errors to hidden neurons which can lead to significant computation and energy consumption on hardware. 
\par
Alternatives such as the \textit{Spike-Train Level Direct Feedback Alignment} (ST-DFA) \citep{st_dfa} or the \textit{Error-Modulated Spike-Timing-Dependent Plasticity} (EMSTDP) \citep{emstdp_dfa_2, emstdp_dfa} algorithms accumulate local changes of weights during the inference and send a single error signal to hidden neurons through random direct feedback connections. These approaches are more energy-efficient because  the error projection is only performed once, thus requiring fewer operations. They have also have been successfully implemented on neuromorphic hardware and demonstrated promising performance and lower energy consumption compared to offline training on GPUs \citep{st_dfa, emstdp_dfa, eprop_dfa_spinnaker}. 
\par
However, they have significant shortcomings. The most important one is that the performance of the algorithms is far behind the state of the art, i.e. far behind BP.  It is generally thought that DFA algorithms  rely on the alignment between weight and feedback connections \citep{random_dfa, align_then_memorise}.  In the context of rate-coded neural networks it has recently  been shown that the strong bias introduced by the random feedback connections affects the convergence rate of DFA. A solution for this was proposed in the form of the FDFA algorithm  \citep{mynnpaperwithbacho}, which enhanced DFA by learning the feedback matrices using Forward-AD. This proved to be  highly successful in the context of rate-coded (non-spiking) neurons. 
\par
The main contribution of this paper is the  \textit{Spiking Forward Direct Feedback Alignment} (SFDFA) algorithm and adaptation of  FDFA to SNNs. While the main idea of the two algorithms are the same, a number of conceptual modifications were necessary for SFDFA. Firstly, Forward-AD is not suitable for SNNs, and we use instead, as an alternative, {\em  graded spikes} to estimate the weight between output and hidden neurons. Secondly, since spikes are non-differentiable, new different approaches are necessary in order  compute gradients. Here we choose an approach based on \citep{fast_deep} which computes gradients of spike times of neurons. Unlike, \citep{fast_deep}, however, we allow multiple spikes per neurons, which in turn required a novel approaches to compute local gradient (see \citep{mypaperwithbacho} for a similar approach). Thirdly, as we found, the approach we took leads to a singularity where the  the gradient diverges; we propose an {\em ad-hoc} solution for this. Using extensive simulations on a variety of benchmark problems, we demonstrate that SFDFA is effective, outperforming previous neuromorphic algorithms on a variety of tasks. However, it remains consistently below the current state of the art, as set by BP.  

\section{Neuron model and notation}

\subsection{Leaky Integrate-and-Fire Neuron}

Ie this work, we use networks of Leaky Integrate-and-Fire (LIF) neurons with the following membrane potential $u_i(t)$ dynamics for the post-synaptic neuron $i$:
\begin{equation} \label{eq:fast_deep_mp}
	\begin{split}
		u_i(t) =& \sum_{j \in \mathcal{P}_i} w_{i,j} \sum_{t_j^z \in \mathcal{T}_j} \Theta\left(t - t_j^z\right) \left[\exp\left(\frac{t_j^z - t}{\tau}\right) - \exp\left(\frac{t_j^z - t}{\tau_s}\right)\right] \\
		& - \vartheta \sum_{t_i^k \in \mathcal{T}_i} \Theta \left(t - t_i^k\right) \exp\left(\frac{t_i^k - t}{\tau}\right)
	\end{split}
\end{equation}
where $t$ is the current time, $\mathcal{P}_i$ is the set of pre-synaptic neurons j connected to the post-synaptic neuron $i$, $t_j^k$ is the time of the $k$-th (postsynaptic) spike of neuron $j$,  and $\mathcal{T}_j:= \{t_j^1, tj^2,\ldots t_j^{n_j}\}$ is the set of timings of all $n_j$ postsynaptic spikes of  neuron $j$, $w_{i,j} \in \mathbb{R}$ is the strength of the synaptic connection between the post-synaptic neuron $i$ and the pre-synaptic neuron $j$, $\tau_s$ and $\tau$ are respectively the synaptic and membrane time constants, $\vartheta \in \mathbb{R}$ is the spiking threshold of the neuron and 
\begin{equation}
	\Theta(x) = \begin{cases}
		1 & \text{ if } x \geq 0 \\
		0 & \text{ otherwise}
	\end{cases}
\end{equation}
is the Heavyside step function that only triggers events that occurred in the past.
\par
In essence, the LIF neuron integrates pre-synaptic spikes over time through its synapses, thus affecting its membrane potential. When the membrane potential $u_i(t)$ reaches the neuron's threshold $\vartheta \in \mathbb{R}$ at time $t$, a post-synaptic spike at time is fired at the output of the neuron and the membrane potential is reset to zero.

\subsection{Exact Gradients of SNNs using Error Backpropagation} \label{seq:bp}

Inspired by its success with DNNs, the error Backpropagation (BP) algorithm has been successfully adapted to SNNs using various approximations that sidestep the non-differentiability aspect and the lack of closed form solution of spikes \citep{spikeprop, slayer, going_deeper_in_snn, spatio_temporal_bp, macro_micro_backpropagation}. More recent approaches have successfully derived exact gradients by using implicit functions describing the relation between membrane potential and spike timings \citep{event_based_exact_gradient_snn}. Alternatively, more recent methods such as Fast \& Deep \citep{fast_deep} and its multi-spike extension introduce constraints to isolate explicit and differentiable closed-form solutions for spike times \citep{supervised_learning_based_on_temporal_coding, gradient_descent_alpha_function, fast_deep}. By fixing  the decay parameters of the membrane potential in eq.~\ref{eq:fast_deep_mp}  to  $\tau=2\tau_s$, a closed form solution for the spike timings $t_i^k$ can be found.
\begin{equation}
\label{spiketiming}
	t_i^x = \tau \ln \left(\frac{2 a_i^x}{b_i^x + \chi_i^x}\right)
\end{equation}
where
\begin{equation}
\label{abequations}
\begin{split}
	a_i^x &:= \sum_{j \in \mathcal{P}_i} w_{ij} \sum_{z=1}^{x}  \exp\left(\frac{t_j^z}{\tau_s}\right) \nonumber\\
	b_i^x &:= \sum_{j \in \mathcal{P}_i} w_{ij} \sum_{z=1}^{x} \exp\left(\frac{t_j^z}{\tau}\right) - \vartheta \sum_{z=1}^{x}   \exp\left(\frac{t_i^z}{\tau}\right)
\end{split}
\end{equation}
and
\begin{equation}
	\chi_i^x := \sqrt{\left(b_i^x\right)^2 - 4 a_i^x \vartheta}
\end{equation}
This explicit expression of post-synaptic spike timings thus allows for exact differentiation --- see \citep{fast_deep} for more details about this method. 
\par
In the context of event-based SNNs, the gradient of given a differentiable loss function $\mathcal{L}$ of a neural network with $L$ layers is defined by all the partial derivatives $\frac{\partial \mathcal{L}}{\partial w_{pq}}$, such as:
\begin{equation}
		 \frac{\partial \mathcal{L}}{\partial w_{pq}}= 
\partial_i\mathcal L^x    {\partial t_i^{x,(L)} \over \partial t_j^{y,(L-1)}} {\partial t_j^{y ,(L-1)}\over \cdots} \ldots {\cdots \over \partial t_k^{z}} \frac{\partial t_k^{z,(l)}}{\partial w_{pq}^{(l)}}
\end{equation}
To simplify the notation, we  introduced the shorthand  $\partial_i\mathcal L^k := \frac{\partial \mathcal{L}}{\partial t_i^{k,(L)}} $ and used  the convention that repeated indices are summed over. Note that the  $t_i^x$ superscripted index runs from $1$ to $n_i$. To simplify notation, we will henceforth omit the layer super-script whenever this can be inferred from context. We will henceforth refer to $\frac{\partial t_i^k}{\partial w_{pq}}$ as the  {\em local derivative} of the spike timing.
\par
Under BP, the update of weights $\Delta^\textrm{BP} w_{pq}$ is then proportional to $\frac{\partial \mathcal{L}}{\partial w_{pq}}$, with the learning rate $\eta$ as the proportionality factor. While this version of BP has yielded good results on SNNs in a number of contexts \citep{}, it has two major drawbacks. \one Updates using BP require information that is non-local to the neuron, which is difficult to achieve on neuromorphic hardware. \two BP requires the hardware to retain a memory of the past states of neurons, which is not compatible with neuromorphic principles. 
\par
A simple solution to this is the above mentioned DFA algorithm, whereby the updates of the weights are not proportional to the gradient of the loss function but are instead made according to the   modified DFA learning rule
\begin{equation}
\label{dfaformula}
\Delta^\textrm{DFA} w_{pq} := \eta  1_x \partial_i\mathcal L^x  b_{ik}^{(l)} \frac{\partial t_k^{z,(l)}}{\partial w_{pq}^{(l)}} 1_z
\end{equation}
Here $b_{ik}^{(l)}$  are the elements of a randomly drawn, but fixed {\em feedback matrix}   $\mathbf{B}^{(l)}$  for layer $l$. The symbol  $1_x$ is an indexed constant of value 1 which implements the sum over the derivatives of the individual spikes. 
\par
DFA sidesteps the need for backpropagating errors backward through space and time and allows for online error computation. In addition, because all post-synaptic spikes receive the same projected error, a local gradient of spikes can be locally computed during the inference without immediately requiring an error signal. For this reason, DFA has been successfully implemented in several neuromorphic hardware \citep{st_dfa, emstdp_dfa, eprop_dfa_spinnaker}.
\par
While efficient to compute, the performance of DFA is typically not competitive with BP. Previously \citep{mynnpaperwithbacho}, we showed in the context of rate-coded neural networks that  the performance can be increased substantially by updating the feedback matrices along with the weights themselves. In order to compute the update to the feedback matrix, we require the  neuron to store an additional value (the {\em grade}, denoted by $d_i$) locally. The grade is initialised to zero, and then updated as follows:
\begin{itemize}
\item
Upon receiving an input spike from neuron $j$, the grade is increased by $w_{i{j_0}} \cdot d_{j_0}$ (note the summation convention does not apply here).
\item
Upon generating  the first spike at time $t_i^0$ a random number is drawn from a normal distribution and stored as $p_i^0$.
\item
Upon generating a spike the grade is reset to zero.  
\end{itemize} 
During training, the feedback matrices are then updated as  follows

As in FDFA, we define an update rule for the feedback connection $b_{o,i}$ that performs an exponential average of the weight estimate, such as:
\begin{equation}
	b_{ji} \leftarrow \left(1 - \alpha\right) b_{ji} + \alpha  d_j  p_i^1
\end{equation}
where $0 < \alpha < 1$ is the feedback learning rate.
\par
In the case of a single hidden layer this learning rule has a clear interpretation. It  is straightforward to show that the  feedback matrices lead  become over time similar to the forward weights $w_{ij}$.  
\begin{equation}
	\begin{split}
		d_o =& \sum_{t_o^k \in \mathcal{T}_o} d_o^k \\
		=& \sum_{i \in \mathcal{P}_o} w_{o,i} \sum_{t_i^z \in \mathcal{T}_i} d_i^z \\
		=& \sum_{i \in \mathcal{P}_o} w_{o,i} \sum_{t_i^k \in \mathcal{T}_i} p_i^k \\
		=& \sum_{i \in \mathcal{P}_o} w_{o,i} \; p_i^1 \\
	\end{split}
\end{equation}
\par
Multiplying $d_o$ by the perturbation $p_i^1$ then leads to an unbiased estimate of the weight $w_{o,i}$.
\begin{equation}
	\begin{split}
		\mathbb{E}\left[d_o \; p_i^1\right] =& \mathbb{E}\left[\sum_{j \in \mathcal{P}_o} w_{o,j} \; p_j^1 \; p_i^1\right] \\
		=& \sum_{j \in \mathcal{P}_o} w_{o,j} \; \mathbb{E}\left[p_j^1\right] \mathbb{E}\left[p_i^1\right] \\
		=& w_{o,i}
	\end{split}
\end{equation}
In that sense, SFDFA then approximates BP. 
\par
%
%
\begin{algorithm}[htbp]
	\caption{Inference of a single hidden neuron $i$ with the Spiking Forward Direct Feedback Alignment algorithm.}
	\label{alg:sfdfa_inference}
	\begin{algorithmic}[1]
		\STATE {\bfseries Input:} The neuron index $i$, the neuron weights $\boldsymbol{w}$, the reset current $\vartheta = \vartheta - u_{\text{rest}}$, the random perturbation $p \sim \mathcal{N}\left(0, 1\right)$ and the set $\mathcal{T}_{\text{pre}}$  of pre-synaptic graded spikes $\left(j, t_j^z, d_j^z\right) \in \mathcal{T}_{\text{pre}}$ sorted in time where $j$ is a pre-synaptic neuron index, $t_j^z$ is a timing and $d_j^z$ is a spike grade.
		\STATE {\bfseries Initialize:} the factors $a=0$ and $b=0$, the spike count $k$, the local gradient $\boldsymbol{\nabla_{\text{loc}} w} = \boldsymbol{0}$, the set of output events $\mathcal{T}_{\text{post}} = \left\{\right\}$, the internal directional derivatives $s = 0$, and the local derivatives synaptic traces $\boldsymbol{f} = \boldsymbol{0}$ and $\boldsymbol{h} = \boldsymbol{0}$.
		\FORALL{$\left(j, t_j^z, d_j^z\right)$ {\bfseries in} $\mathcal{T}_{\text{pre}}$}
		\STATE $f_{j} \leftarrow f_{j} \exp\left(\frac{t_j^z}{\tau_s}\right)$ \COMMENT{Update local derivative traces}
		\STATE $h_{j} \leftarrow h_{j} \exp\left(\frac{t_j^z}{\tau}\right)$
		\STATE $a \leftarrow a + w_{j} \exp\left(\frac{t_j^z}{\tau_s}\right)$ \COMMENT{Integrate pre-synaptic spike into quadratic factors}
		\STATE $b \leftarrow b + w_{j} \exp\left(\frac{t_j^z}{\tau}\right)$
		\STATE $s \leftarrow s + w_{j} \; d_j^z$ \COMMENT{Update internal directional derivative}
		\IF{the neuron fires a valid spike}
		\STATE {\bfseries continue}
		\ENDIF
		\STATE $t \leftarrow \tau \ln \left(\frac{2a}{b+\sqrt{b^2 - 4a\vartheta}}\right)$ \COMMENT{Compute post-synaptic spike timing}
		\STATE $\boldsymbol{t^\prime} \leftarrow - \frac{\tau}{a_i^k \exp\left(\frac{-t}{\tau_s}\right)} \left[\boldsymbol{h} \exp\left(\frac{-t}{\tau}\right) - \boldsymbol{f} \exp\left(\frac{-t}{\tau_s}\right)\right]$ \COMMENT{Compute spike derivative}
		\STATE $\boldsymbol{\nabla_{\text{loc}} w} \leftarrow \boldsymbol{\nabla_{\text{loc}} w} + \boldsymbol{t^\prime}$ \COMMENT{Update local gradient}
		\STATE $\boldsymbol{h} \leftarrow \boldsymbol{h} - \frac{\vartheta}{\tau} \exp\left(\frac{t}{\tau}\right) \boldsymbol{t^\prime}$ \COMMENT{Apply recurrence}
		\STATE $b \leftarrow b - \vartheta \exp\left(\frac{t}{\tau}\right)$ \COMMENT{Reset of membrane potential}
	\STATE $d \leftarrow s + p \text{ \bfseries if } k=0 \text{ \bfseries else } s$ \COMMENT{Compute directional derivative as spike grade}
		\STATE add $\left(i, t, d\right)$ to $\mathcal{T}_{\text{post}}$
		\ENDFOR
		\RETURN{$\mathcal{T}_{\text{post}}$}, $\boldsymbol{\nabla_{\text{loc}} w}$
	\end{algorithmic}
\end{algorithm}
\begin{algorithm}[htbp]
	\caption{Weights update of a single hidden neuron $i$ with the Spiking Forward Direct Feedback Alignment algorithm.}
	\label{alg:sfdfa_update}
	\begin{algorithmic}[1]
		\STATE {\bfseries Input:} the neuron weights $\boldsymbol{w}$, the local gradient $\boldsymbol{\nabla_{\text{loc}} w}$, the feedback connections $\boldsymbol{b}$, the random perturbation $p$, the output directional derivatives $\boldsymbol{d}$, the output errors $\boldsymbol{\delta}$ and the learning rates $\lambda$ and $\alpha$.
		
		\STATE $\boldsymbol{w} \leftarrow \boldsymbol{w} - \lambda \; \boldsymbol{\delta} \boldsymbol{b} \boldsymbol{\nabla_{\text{loc}} w}$ \COMMENT{Update weights}
		\STATE $\boldsymbol{b} \leftarrow \left(1-\alpha\right) \boldsymbol{b} + \alpha \; p \boldsymbol{d}$ \COMMENT{Update feedback connections}
		\RETURN{$\boldsymbol{w}$}, $\boldsymbol{b}$
	\end{algorithmic}
\end{algorithm}

\subsection{Computing the DFA update rule}

In this section, we now describe how we implemented the SFDA learning rule in practice in the context of this paper. 
Many definitions exist for these local gradients of spikes. For example, the gradient of the membrane potential at spike times \citep{st_dfa} or surrogate local gradients \citep{eprop_dfa} can be used. Other works, such as EMSTDP \citep{emstdp_dfa}, use biologically-plausible local gradients like STDP or simple accumulations of the number of pre-synaptic spikes as in eRBP \citep{erbp_dfa}. Here we show that within our framework we can explicitly compute  the errors $\partial_i\mathcal L^x$  and the of the local gradient $ \frac{\partial t_k^{z,(l)}}{\partial w_{pq}^{(l)}}$ in eq.~\ref{dfaformula}. In the following two subsections we will show how to do this. 

\subsubsection{Computing the error}

In the particular examples that we consider here, we will use a spike count $n$ to decode the output, with a  loss function defined as:
\begin{equation}
\mathbf L(n) := \sum_i \left(   y_i - n_i  \right)^2
\end{equation}
Here, $y_i$ is a particular training example  and $n_i$ the actual spike count of the output neuron of the spiking network; the sum runs over a mini-batch. 
\par
The spike count $n$ is a complicated function of the spike timings of the output neurons, and consequently the spike timings of hidden neurons as well, as per eq.~\ref{dfaformula}. We can then see that for this particular choice the error term in eq.~\ref{dfaformula} can be split up as follows.
\begin{equation}
1_x \partial_i\mathcal L^x =  {\partial \mathcal L\over \partial n} 1_x {\partial n\over \partial t^x_i}
\end{equation}
Computing the partial derivatives with respect to spike timings is computationally expensive. To simplify the computation, we absorb those into the random feedback matrices, which gives the following simplified DFA update rule:
\begin{equation}
\label{dfaformulasimplified}
\Delta^\textrm{DFA} w_{pq} := \eta   {\partial \mathcal L\over \partial n} b_{k}^{(l)} \frac{\partial t_k^{z,(l)}}{\partial w_{pq}^{(l)}} 1_z
\end{equation}
Here, again, we used the notation that repeated indices are summed over.

\subsubsection{Computing the local gradient} 

Next, we describe how to compute the local gradient in eq.~\ref{dfaformula}. We observe that in eq.~\ref{spiketiming} $t_i^x$ depends explicitly on $w_{ij}$, and also depends on $t_i^{x-1}$, which in turn depends on $w_{ij}$ and its preceding spikes (cf.~eq.~\ref{abequations}). In order to capture the gradient accurately, we therefore need to replace the partial derivative in eq.~\ref{dfaformula} by the  total derivative of the spike timing with respect to the weight $w_{pq}$. To compute this, we use the general formula for the total derivative as an {\em ansatz}.
\begin{equation}
\abl{t_i^x}{w_{ij}} = \frac{\partial t_i^x}{\partial w_{ij}} + 
{\partial t_i^x \over \partial t_i^{x-1}}    \abl{t_i^{x-1}}{w_{ij}} + \cdots +   
{\partial t_i^x \over \partial t_i^{1}}    \abl{t_i^{1}}{w_{ij}} 
\end{equation}
Substituting in the explicit expressions for $t_i^x$ as given by eq.~\ref{spiketiming} and evaluating the derivatives,  we  find after some straightforward, albeit tedious algebra an expression for the total derivative of the spike timing,
\begin{equation}
\label{localgradient}
	\abl{t_i^x}{w_{ij}}	= \frac{\tau}{a_i^x} \left[1 +  \frac{\vartheta}{\chi_i^x} \exp\left(\frac{t_i^x}{\tau}\right)\right] f_{ij}^x - \frac{\tau}{\chi_i^x} \; h_{ij}^x,
\end{equation}
where we used  the shorthands
\begin{equation}
\label{feq}
	f_{ij}^x = \sum_{\{z: t_j^z\leq t_i^x\}}  \exp\left(\frac{2 t_j^z}{\tau}\right)
\end{equation}
and
\begin{equation}
\label{heq}
	h_{ij}^x =   \sum_{\{z: t_j^z\leq t_i^x\}}  \exp\left(\frac{t_j^z}{\tau}\right)
  -\sum_{z=1}^x  \left(    \frac{\vartheta}{\tau} \exp\left(\frac{t_i^z}{\tau}\right)  \abl{t_i^z}{w_{ij}}\right)
\end{equation}
and used the condition that $\tau=2\tau_s$.

In appendix \ref{simplified}, we derive a mathematically equivalent, but computationally more convenient and faster form of this equation.
\begin{equation} \label{eq:loc_grad_final}
	\abl{t_i^x}{w_{ij}}	= 	 \frac{\tau}{a_i^x \exp\left(\frac{-2t_i^x}{\tau}\right) - \vartheta} \left[f_{ij}^x \exp\left(\frac{-2t_i^x}{\tau}\right) - h_{ij}^x \exp\left(\frac{-t_i^x}{\tau}\right)\right]
\end{equation}

\subsection{Critical Points of Local Gradients}

\begin{figure}[htbp]
	\centering
		\includegraphics[width=0.5\textwidth]{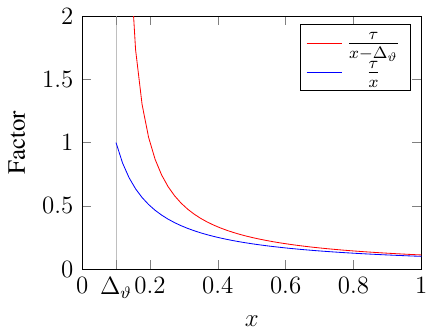}
	\caption{Original factor $\frac{\tau}{x - \vartheta}$ (red line) and the modified factor $\frac{\tau}{x}$ as a function of $x=a_i^k \exp\left(\frac{-t_i^k}{\tau_s}\right)$. As $a_i^k \exp\left(\frac{-t_i^k}{\tau_s}\right)$ approaches $\vartheta$, the original factor diverges towards infinity while the modified factor is bounded.}
	\label{fig:factors}
\end{figure}

It is apparent from eq.~\ref{localgradient} that the local gradient diverges when $\chi \to 0$ or equivalently when $a_i^x \exp\left(\frac{-t_i^x}{\tau_s}\right)= \vartheta$ (cf.~\citep{event_based_exact_gradient_snn, spikeprop_hair_trigger, spikeprop_surge} for similar observation).  A common solution to mitigate gradient explosion is adding additional mechanisms to limit the size of the gradient such as \textit{gradient clipping} \citep{event_based_exact_gradient_snn, spikeprop_grad_clipping}. However, gradient clipping requires computing the norm of the gradient, which cannot be locally performed on neuromorphic hardware. In this work, we propose a simple {\em ad-hoc} modification of the local gradient to remove the cause of the critical points.  
\par
Our solution is to  drop $\vartheta$ from eq.~\ref{eq:loc_grad_final},  thus obtaining:
\begin{equation} \label{eq:loc_grad_fixed}
		\abl{t_i^x}{w_{ij}} = \frac{\tau}{a_i^x \exp\left(\frac{-t_i^x}{\tau_s}\right)} \left[f_{ij}^x \exp\left(\frac{-t_i^x}{\tau_s}\right) - h_{ij}^x \exp\left(\frac{-t_i^x}{\tau}\right)\right]
\end{equation}
Here, $\frac{\tau}{a_i^k \exp\left(\frac{-t_i^k}{\tau_s}\right)} \leq \frac{\tau}{\vartheta}$ has an upper bound and does not diverge towards infinity (see Figure \ref{fig:factors}).
While this creates an additional bias in the gradient estimates, dropping the term $\vartheta$ avoids exploding gradients  and  improves the stability of the weights updates.

\subsection{Neuromorphic Hardware Compatibility} \label{sec:sfdfa_hardware_compatibility}

The local gradient in Equation \ref{eq:loc_grad_fixed} is adapted for computation on von Neumann computers. Therefore, to make the SFDFA algorithm compatible with neuromorphic hardware, we derive an eligibility trace that can be evaluated with a system of ordinary differential equations on hardware.
\par
Eq.~\ref{eq:loc_grad_fixed} can be written as:
\begin{equation}
		\abl{t_i^x}{w_{ij}} = - \frac{\tau}{I_i(t_i^x)} e_{ij}(t_i^x)
\end{equation}
where
\begin{equation}
	\begin{split}
		I_i(t) =& a_i^x \exp\left(\frac{-t}{\tau_s}\right) \\ =& \sum_{j \in \mathcal{P}_i} w_{ij} \sum_{\{z: t^z_j \leq t\}}  \alpha\left(t - t_j^z\right)
	\end{split}
\end{equation}
is the synaptic current at the spike time, which is computed by the LIF neuron \citep{spiking_neuron_models}, and
\begin{equation} \label{eq:sfdfa_eligibility_trace_srm}
	\begin{split}
		e_{ij}(t) =& \sum_{ \{z: t^z_j \leq t\}}  \underbrace{\left[\exp\left(\frac{t_j^z - t}{\tau}\right) -  \exp\left(\frac{t_j^z - t}{\tau_s}\right)\right]}_{=\epsilon\left(t - t_j^z\right)} \\
		& - \sum_{  \{z: t^z_j \leq t\} }  \underbrace{\vartheta \exp\left(\frac{t_i^z - t}{\tau}\right)}_{=\eta\left(t - t_i^z\right)}  \abl{t_i^x}{w_{ij}}\frac{1    }{\tau}
	\end{split}
\end{equation}
is an eligibility trace computed over time that coincides with the desired expression required by the local gradient at the spike time $t_i^k$. We can observe that Equation \ref{eq:sfdfa_eligibility_trace_srm} has a form that is similar to the mapping of the LIF neuron to the SRM model.  Therefore, this expression can be implemented on neuromorphic hardware using a LIF model at each synapse and locally be used at post-synaptic spike times for the computation of local gradients.

\section{Empirical Results} \label{sec:sfdfa_results}

We now present our empirical results with our proposed SFDFA algorithm. We empirically highlight the critical points in the exact local gradient of spikes and demonstrate the effectiveness of our modified local gradient in preventing gradient explosions. We then compare the performance and convergence rate of SFDFA with DFA and BP on several benchmark datasets as well as the weight and gradient alignment of DFA and SFDFA.

\subsection{Critical Points Analysis}

\begin{figure}[t]
	\centering
		\centering
	\subfloat[Exact local gradient]{\includegraphics{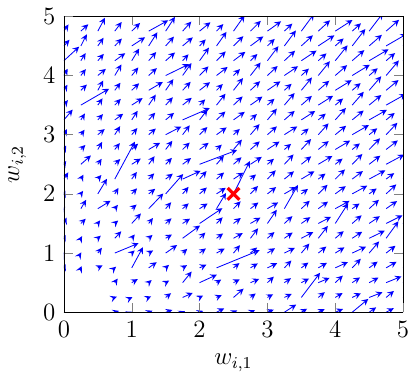}	\label{fig:grad_field_original}}
	\subfloat[Modified local gradient]{\includegraphics{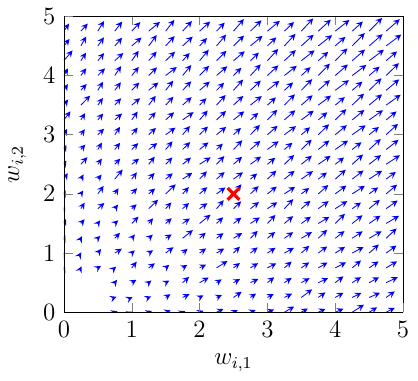} \label{fig:grad_field_modified}}
	\caption{Gradient fields of a single neuron  with two inputs computed using the exact local gradients \protect\subref{fig:grad_field_original} and the modified local gradients \protect\subref{fig:grad_field_modified}. We can observe that the exact local gradient contains critical points where the norm is abnormally large compared to neighboring regions. However, with the modified local gradients, the norm of the gradient is consistent throughout the weight space, mitigating the gradient explosion caused by the critical points. The red crosses at  correspond to the critical point visualized in Figure \protect\ref{fig:instability}. See main text for details on how this figure was generated. For these simulations, we used $\tau_s=0.01$, $\theta=0.01$ and the time window was of length $0.01$ seconds. }
	\label{fig:grad_fields}
\end{figure}

We now analyze the stability of the exact local gradient of spikes and compare it with our modified local gradient, defined in Equation \ref{eq:loc_grad_fixed}. To gain an intuitive understanding of the problem, we first consider a toy example of  a two-input neuron receiving four pre-synaptic spikes. We  evaluate the local gradient with the exact derivative of spikes (Equation \ref{eq:loc_grad_final}) and with our modified local gradient (Equation \ref{eq:loc_grad_fixed}). 
Figure \ref{fig:grad_fields} shows the respective gradient fields computed using the exact and modified local gradients. It can be observed in Figure \ref{fig:grad_field_original} that the exact local gradient contains several critical points where its norm is significantly larger than neighboring vectors. 
\par
\begin{figure}[H]
	\centering 
		\subfloat[Membrane potential]{\includegraphics[width=0.45\textwidth]{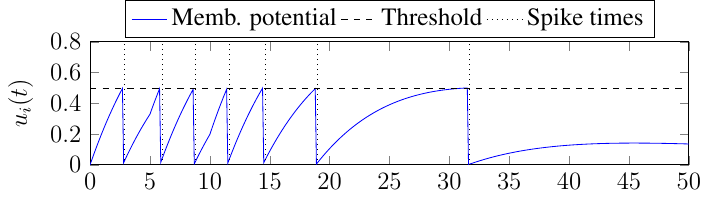} \label{fig:instability_mp}}
		\subfloat[Input current]{\includegraphics[width=0.45\textwidth]{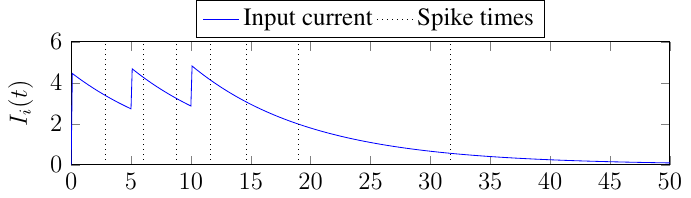} \label{fig:instability_current}}\\
		\subfloat[Factors]{\includegraphics[width=0.45\textwidth]{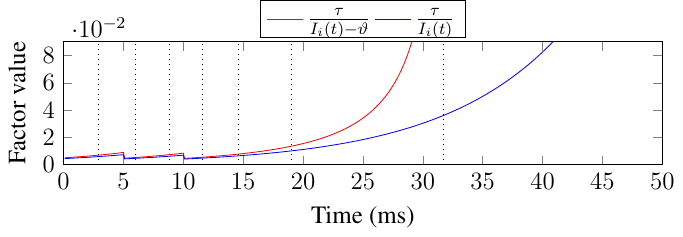} \label{fig:instability_factors}}
		\subfloat[Local gradient]{\includegraphics[width=0.45\textwidth]{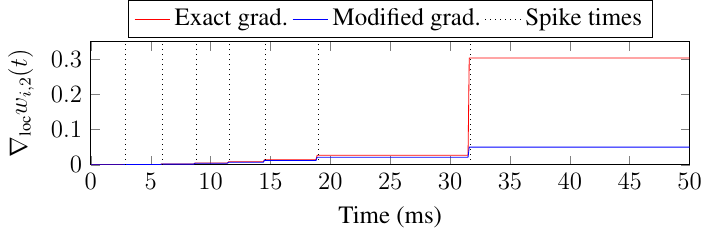} \label{fig:instability_loc_gradient}}
	\caption{Evolution over time of the membrane potential \protect\subref{fig:instability_mp}, input current \protect\subref{fig:instability_current}, gradient factors \protect\subref{fig:instability_factors} and local gradients \protect\subref{fig:instability_loc_gradient} at an instability point ($w_{i,1}=2.5$ and $w_{i,2}=2.0$ in Figure \ref{fig:grad_fields}) in a minimal example of a single spiking neuron with two inputs. Here  all vertical dotted lines correspond to post-synaptic spike times. The red lines in both  \protect\subref{fig:instability_factors} and \protect\subref{fig:instability_loc_gradient} represent the original factor and local gradient, while the blue lines represent the modified factor and local gradient. These figures show that the last post-synaptic spike fired by the neuron occurs when the membrane potential narrowly reaches the threshold. This narrow threshold crossing is attributed to the low input current $I_i(t)$ which causes large factors $\frac{\tau}{I_i(t) - \vartheta}$ and consequently large local gradients. In contrast, our modified factor $\frac{\tau}{I_i(t)}$ restricts the amplitude of the factor, thereby moderating the scale of the gradient. For these simulations, we used $\tau_s=0.01$, $\theta=0.01$ and the time window was of length $0.01$ seconds. }
	\label{fig:instability}
\end{figure}

To investigate the cause behind these large norms, we examined the neuron's internal state over time at one of these critical points (marked in red in Figure \ref{fig:grad_fields}). Figure \ref{fig:instability} depicts the temporal evolution of the membrane potential, input current, factor $\frac{\tau}{I_i(t) - \vartheta}$, and the computed local gradient. Notably, it can be observed that the last post-synaptic spike fired by the neuron narrowly crosses the threshold (i.e. hair trigger) due to a low input current. This low current leads to the factor $\frac{\tau}{I_i(t) - \vartheta}$ becoming large due to its divergence when $I_i(t)$ approaches $\vartheta$ (see Figure \ref{fig:factors}). Consequently, the local gradient at this spike time significantly increases.
\par
On the other hand, Figure \ref{fig:grad_fields} demonstrates that our proposed modified local gradient maintains a consistent norm across the weight space without any instability points where the norm deviates abnormally from neighboring vectors. Additionally, Figure \ref{fig:instability} illustrates that, at the spike time when the membrane potential narrowly reaches the threshold, the modified factor $\frac{\tau}{I_i(t)}$ exhibits a significantly lower value compared to the original factor $\frac{\tau}{I_i(t) - \vartheta}$. This is because the modified factor is upper-bounded, as shown in Figure \ref{fig:factors}. As a result, the contribution of this spike to the modified local gradient is substantially reduced in comparison to its contribution to the exact local gradient.
\par
To assess the impact of critical points on the convergence of SNNs, we conducted an experiment using the MNIST dataset. Identical SNNs were trained, both containing two layers of fully connected neurons and sharing the same hyperparameters. These networks were initialized with identical weights, differing only in the type of local gradient used for training. One network was trained with the exact local gradient, while the second network was trained with the modified gradient defined in Equation \ref{eq:loc_grad_fixed}.
\par
\begin{figure}[H]
			\subfloat[Train accuracy]{\includegraphics[width=0.45\textwidth]{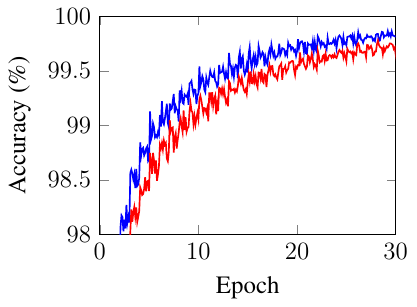}\label{fig:factor_mnist_train_acc}}
			\subfloat[Test accuracy]{\includegraphics[width=0.45\textwidth]{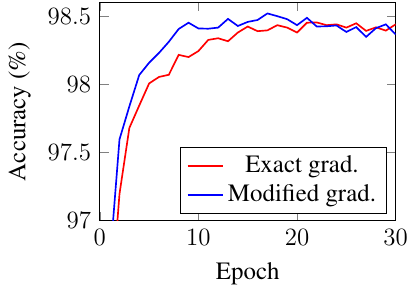}\label{fig:factor_mnist_test_acc}}\\
			\subfloat[Train loss]{\includegraphics[width=0.45\textwidth]{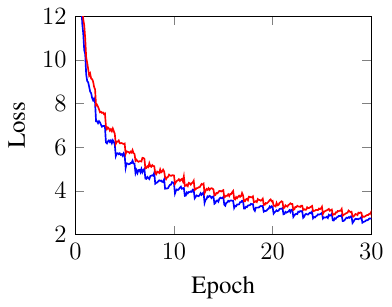}\label{fig:factor_mnist_train_loss}}
			\subfloat[Test loss]{\includegraphics[width=0.45\textwidth]{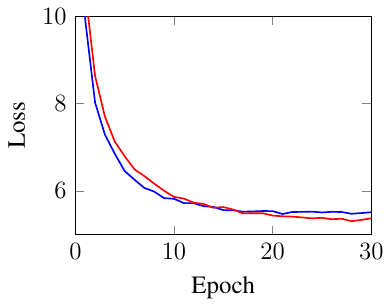}\label{fig:factor_mnist_test_loss}}
	\caption{Evolution of the training accuracy (Figure \ref{fig:factor_mnist_train_acc}), test accuracy (Figure \ref{fig:factor_mnist_test_acc}), train loss (Figure \ref{fig:factor_mnist_train_loss}) and test loss (Figure \ref{fig:factor_mnist_test_loss}) during the training of a two-layers SNN on the MNIST dataset. Red lines correspond to the metrics of the SNN updated using the exact local gradient while blue lines correspond to the metrics of the SNN updated using the modified local gradient defined in Equation \ref{eq:loc_grad_fixed}. These figures show that the modified gradient converges slightly faster than the exact local gradient.}
	\label{fig:factor_mnist}
\end{figure}

Figure \ref{fig:factor_mnist} shows the evolution of the training loss, training accuracy, test loss, and test accuracy for each network during the training of each network. It can notably be observed that the network trained with the modified local gradient converges slightly faster than the SNN trained with the exact local gradient. This suggests that the modified gradient mitigates the impact of instabilities on convergence, leading to enhanced learning.
 
\subsection{Performance}

We evaluate the performance of the proposed SFDFA algorithm as well as BP and DFA in SNNs with four benchmark datasets, namely MNIST \citep{mnist}, EMNIST \citep{emnist}, Fashion MNIST \citep{fashion_mnist} and the {\em Spiking Heidelberg Digits} (SHD) dataset \citep{heidelberg_dataset}.  To encode the pixel values of static images (e.g. MNIST, EMNIST, and Fashion MNIST) into spikes that can be used as inputs of SNNs, we applied a  coding scheme with an encoding window of T = 100 ms. To do this, we normalised the pixel values and calculated the spike time $t$ for a pixel value $x$ as
\begin{equation}
t=T(1-x)
\end{equation}
where $T$ is the length of encoding window, which was chosen to be 100ms throughout.  For pixel values of 0 (i.e. black), no spikes were generated. In this way, TTFS therefore produces sparse temporal encoding of greyscale images that are fast to process in my event-based simulator. For the SHD dataset, I used the spike trains as provided by the dataset.
\par
We used a spike count strategy to decode the output spike counts of the SNNs.  Specifically, each output neuron of the network encoded a possible category for the data. The target spike count for the correct category was set to 10, whereas the target spike count for the incorrect category was set to 3. Note that we did not choose a target of 0 for the latter to avoid dead neurons.   We then used a spike count \textit{Mean Squared Error} loss function for training.
We benchmarked our method with fully-connected SNNs of different sizes. For MNIST and EMNIST, we used two-layer SNNs with 800 hidden neurons, for Fashion MNIST, a three-layer SNN with 400 hidden neurons per hidden layer, and a two-layer SNN with 128 hidden neurons for the SHD dataset. We also compared it with BP applied to an SNN, as described in \cite{mypaperwithbacho},
\par
Table \ref{table:sfdfa_performances} summarizes the performance of each method on each dataset. We can observe that both the DFA and SFDFA algorithms achieve test performances similar to that of BP for the MNIST, EMNIST, and Fashion MNIST datasets. However, for the SHD dataset, neither algorithm attains a level of accuracy comparable to BP. This indicates that the direct feedback learning approach with a single error signal fails to effectively generalize when applied to temporal data.
\begin{table}[htbp]
	\centering
	\caption{Average best test performance of BP, DFA and the proposed SFDFA on the MNIST, EMNIST, Fashion MNIST and SHD datasets.}
	\label{table:sfdfa_performances}
	
	\begin{tabular}{|c|cccc|}
		\hline
		Dataset & Architeture & BP & DFA & SFDFA \\
		\hline
		MNIST & 800-10 & 98.88 $\pm$ 0.02\% & 98.42 $\pm$ 0.06\% & 98.56 $\pm$ 0.04\% \\
		EMNIST & 800-47 & 85.75 $\pm$ 0.06\% & 79.48 $\pm$ 0.11\% & 82.33 $\pm$ 0.10\% \\
		Fashion MNIST & 400-400-10 & 90.19 $\pm$ 0.12\% & 89.41 $\pm$ 0.12\% & 89.73 $\pm$ 0.17\% \\
		SHD & 128-20 & 66.79 $\pm$ 0.66\% & 52.70 $\pm$ 2.30\% & 54.63 $\pm$ 1.16\% \\ \hline
	\end{tabular}
\end{table}
\par
Despite this limitation, the proposed SFDFA algorithm consistently outperforms DFA across all benchmarked datasets. However, there remains a noticeable gap between the performance of SFDFA and BP, particularly as the complexity of the task increases. 

\subsection{Convergence}

The FDFA algorithm introduced in \citep{mynnpaperwithbacho} demonstrated improvements in terms of convergence compared to DFA. To evaluate if this is also the case in SNNs, we recorded the train loss, train accuracy, test loss, and test accuracy of SNNs trained with the DFA and SFDFA algorithms on each dataset.

\begin{figure}[H]
			\subfloat[Test accuracy]{\includegraphics[width=0.45\textwidth]{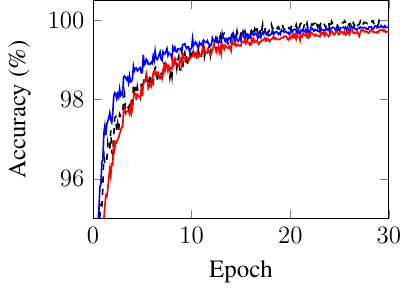}\label{fig:sfdfa_mnist_train_acc}}
			\subfloat[Test accuracy]{\includegraphics[width=0.45\textwidth]{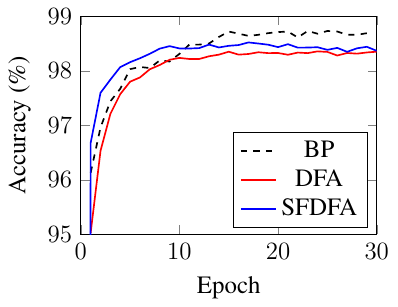}\label{fig:sfdfa_mnist_test_acc} }\\
			\subfloat[Train loss]{\includegraphics[width=0.45\textwidth]{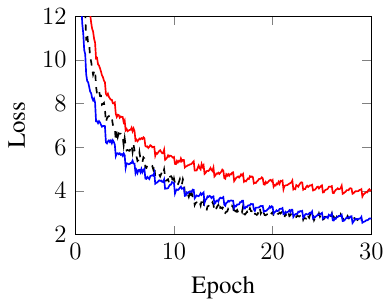}\label{fig:sfdfa_mnist_train_loss}}
			\subfloat[Test loss]{\includegraphics[width=0.45\textwidth]{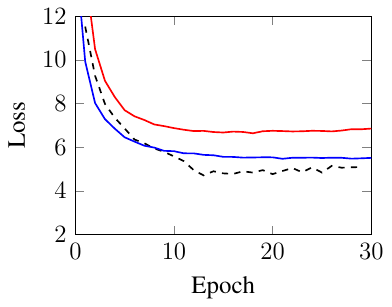}\label{fig:sfdfa_mnist_test_loss}}
	\caption{Evolution of the averaged train accuracy (Figure \ref{fig:sfdfa_mnist_train_acc}), test accuracy (Figure \ref{fig:sfdfa_mnist_test_acc}), train loss (Figure \ref{fig:sfdfa_mnist_train_loss}) and test loss (Figure \ref{fig:sfdfa_mnist_test_loss}) during the training of two-layers SNNs on the MNIST dataset. Black dashed lines correspond to BP. Blue and red solid lines correspond to the SFDFA and DFA algorithms respectively.}
	\label{fig:sfdfa_mnist}
\end{figure}

\begin{figure}[H]
			\subfloat[Train accuracy]{\includegraphics[width=0.45\textwidth]{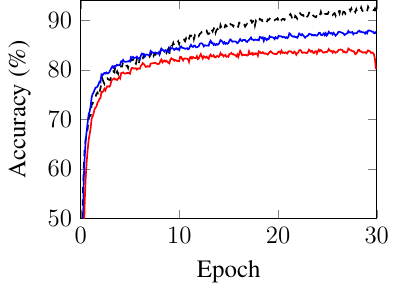}\label{fig:fdfa_emnist_train_acc}}
			\subfloat[Test accuracy]{\includegraphics[width=0.45\textwidth]{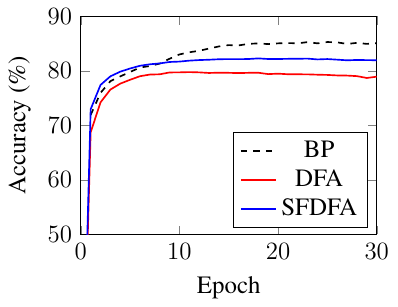}\label{fig:fdfa_emnist_test_acc}}\\
			\subfloat[Train loss]{\includegraphics[width=0.45\textwidth]{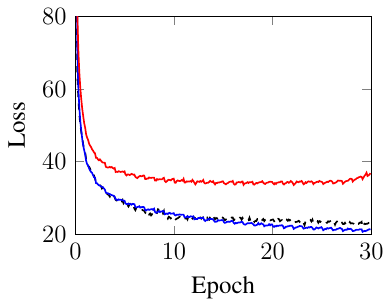}\label{fig:fdfa_emnist_train_loss}}
			\subfloat[Test loss]{\includegraphics[width=0.45\textwidth]{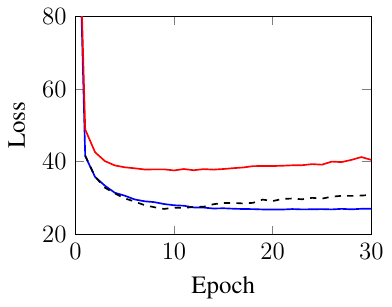}\label{fig:fdfa_emnist_test_loss}}
	\caption{Evolution of the averaged train accuracy \protect\subref{fig:sfdfa_mnist_train_acc}, test accuracy \protect\subref{fig:sfdfa_mnist_test_acc}, train loss (Figure \ref{fig:sfdfa_mnist_train_loss}) and test loss (Figure \ref{fig:sfdfa_mnist_test_loss}) during the training of two-layers SNNs on the EMNIST dataset. Black dashed lines correspond to BP. Blue and red solid lines correspond to the SFDFA and DFA algorithms respectively.}
	\label{fig:sfdfa_emnist}
\end{figure}

\begin{figure}[H]
			\subfloat[Train accuracy]{\includegraphics[width=0.45\textwidth]{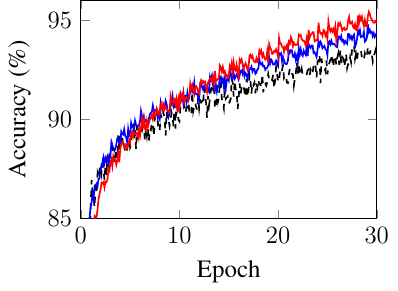}\label{fig:fdfa_fashion_mnist_train_acc}}
			\subfloat[Test accuracy]{\includegraphics[width=0.45\textwidth]{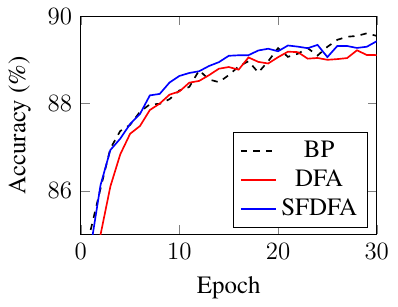}\label{fig:fdfa_fashion_mnist_test_acc}}\\
			\subfloat[Train loss]{\includegraphics[width=0.45\textwidth]{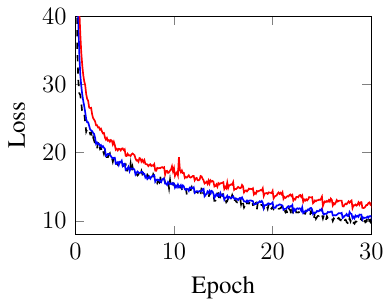}\label{fig:fdfa_fashion_mnist_train_loss}}
			\subfloat[Test loss]{\includegraphics[width=0.45\textwidth]{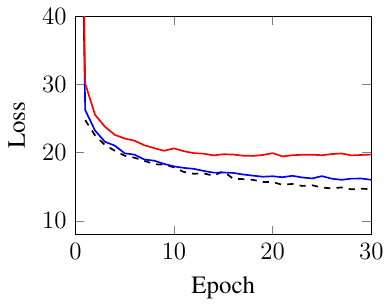}\label{fig:fdfa_fashion_mnist_test_loss}}
	\caption{Evolution of the averaged train accuracy \protect\subref{fig:sfdfa_mnist_train_acc}, test accuracy \protect\subref{fig:sfdfa_mnist_test_acc}, train loss (Figure \ref{fig:sfdfa_mnist_train_loss}) and test loss \protect\subref{fig:sfdfa_mnist_test_loss} during the training of two-layers SNNs on the Fashion MNIST dataset. Black dashed lines correspond to BP. Blue and red solid lines correspond to the SFDFA and DFA algorithms respectively.}
	\label{fig:sfdfa_fashion_mnist}
\end{figure}

\begin{figure}[H]
	\centering
			\subfloat[Train accuracy]{\includegraphics[width=0.45\textwidth]{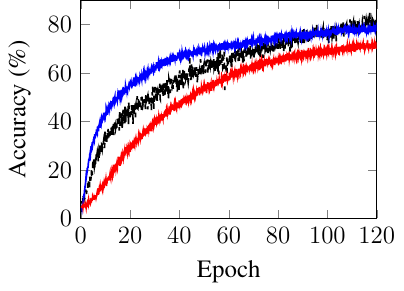}\label{fig:fdfa_shd_train_acc}}
			\subfloat[Test accuracy]{\includegraphics[width=0.45\textwidth]{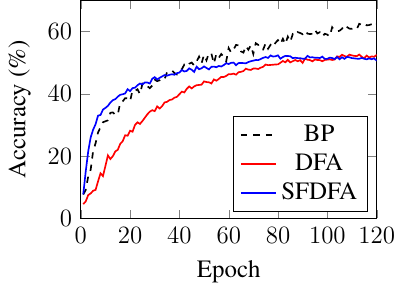}\label{fig:fdfa_shd_test_acc}}\\
			\subfloat[Train loss]{\includegraphics[width=0.45\textwidth]{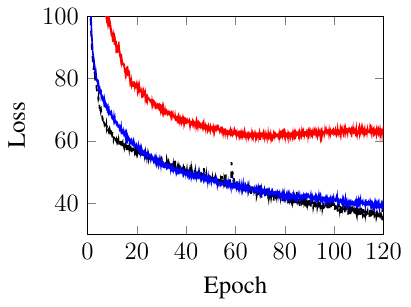}\label{fig:fdfa_shd_train_loss}}
			\subfloat[Test loss]{\includegraphics[width=0.45\textwidth]{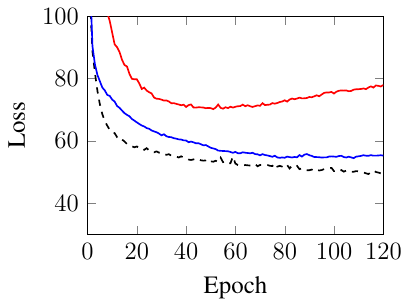}\label{fig:fdfa_shd_test_loss}}
	\caption{Evolution of the averaged train accuracy \protect\subref{fig:sfdfa_mnist_train_acc}, test accuracy \protect\subref{fig:sfdfa_mnist_test_acc}, train loss \protect\subref{fig:sfdfa_mnist_train_loss} and test loss \protect\subref{fig:sfdfa_mnist_test_loss} during the training of two-layers SNNs on the Spiking Heidelberg Digits dataset. Black dashed lines correspond to BP. Blue and red solid lines correspond to the SFDFA and DFA algorithms respectively.}
	\label{fig:sfdfa_shd}
\end{figure}
\par
Figures \ref{fig:sfdfa_mnist}, \ref{fig:sfdfa_emnist}, \ref{fig:sfdfa_fashion_mnist} and \ref{fig:sfdfa_shd} show the evolution of the different metrics with the MNIST, EMNIST, Fashion MNIST and SHD datasets respectively. It can be observed that a gap exists in both the train and test loss between the BP and DFA algorithms. In contrast, the train and test loss for the proposed SFDFA algorithm closely follow the loss values of BP, indicating similar convergence rates. Interestingly, during the initial stage of training, both the train and test accuracies of the SFDFA algorithm increase faster than both BP and DFA. This behavior is particularly prominent in Figures \ref{fig:sfdfa_mnist} and \ref{fig:sfdfa_shd}. However, in the later stage of training, the test accuracy of SFDFA plateaus while the test accuracy of BP continues to improve. This suggests that the proposed SFDFA algorithm does not generalize as much as BP on test data. However, compared to DFA, the test accuracy of the SFDFA algorithm increases significantly faster, especially on temporal data (see Figure \ref{fig:sfdfa_shd}). Altogether, we find that the SFDFA algorithm shows better  convergence than DFA, especially for more difficult tasks.

\subsection{Weights and Gradient Alignment}

\begin{figure}[t]
			\centering
			\subfloat[DFA]{\includegraphics[width=0.45\textwidth]{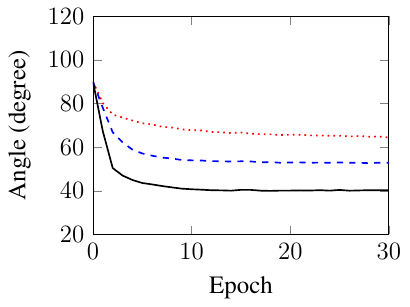}\label{fig:grad_alignment_dfa}}
			\subfloat[SFDFA]{\includegraphics[width=0.45\textwidth]{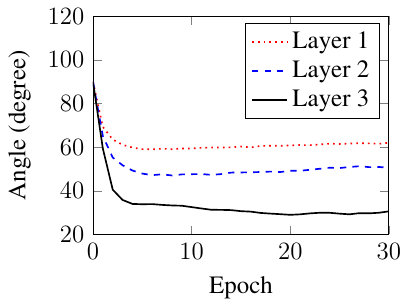}\label{fig:grad_alignment_sfdfa}}
	\caption{Layerwise alignment between spiking gradient estimates and the true gradient computed using BP. These figures show that the SFDFA algorithm Figure \protect\subref{fig:grad_alignment_sfdfa} aligns better with the true gradient than DFA \protect\subref{fig:grad_alignment_dfa}, especially in layers close to the outputs.}
	\label{fig:sfdfa_gradient_alignment}
\end{figure}

\begin{figure}[t]
	\centering
	\includegraphics{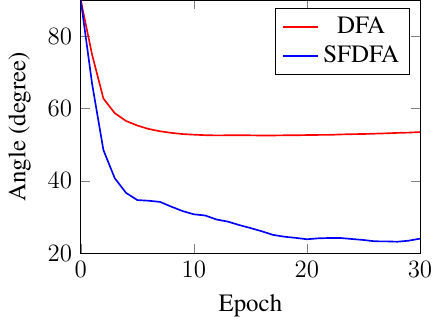}
	\caption{Evolution of the alignment between the output weights and last hidden layer feedbacks in a 4-layer SNN trained on MNIST with DFA (red line) and SFDFA (blue line). The weights and feedback connections align faster and better with the SFDFA algorithm than with DFA. Moreover, the weights alignment in SNNs correlates with the gradient alignment (see Figure \ref{fig:sfdfa_gradient_alignment}).}
	\label{fig:sfdfa_weight_alignment}
\end{figure}

We measured the bias of the gradient estimates by recording the layer-wise alignment between the approximate gradients and the true gradients computed using BP (see \citep{mynnpaperwithbacho}). For this experiment, we trained a 4-layer SNN on the MNIST dataset for 30 epochs using both the DFA and SFDFA algorithms to compute gradient estimates. We then calculated the angles between these estimates and the true gradient computed by BP. 
\par
The evolution of the resulting angles during training is given in Figure \ref{fig:sfdfa_gradient_alignment}. It can be observed that the gradient estimates provided by FDFA align faster and exhibit a lower angle with respect to BP when compared with DFA. This indicates that the proposed SFDFA algorithm achieves a lower level of bias earlier than DFA, with weight updates that better follow BP. However, it is important to note that the gap in alignment between SFDFA and DFA diminishes with the depth of the layer. Specifically, when comparing the alignments of the third layer (i.e. closest to the output) trained with SFDFA and DFA, the former demonstrates significantly better alignment. However, in the case of the first layer, both methods exhibit similar levels of alignment. This suggests that the benefits of SFDFA become more pronounced in layers close to the outputs.
\par
In contrast with the FDFA algorithm which estimates derivatives, the SFDFA algorithm approximates the derivatives of spikes by estimating the weights connections between hidden and output neurons. Therefore, in addition to the layer-wise gradient alignment, we measured the angle between the vectors represented by the output weights and the vector represented by the feedback connections for the last hidden layer. Also known as \textit{weight alignment}, it is believed to be the main source of gradient alignment in DFA \citep{align_then_memorise}. Figure \ref{fig:sfdfa_weight_alignment} shows the evolution of the alignment between the flattened output weights and the flattened feedback connections to the last hidden layer. The alignments for deeper hidden layers were ignored as the exact forward weights for these layers are unknown. We can see in this figure that the output weight and the feedback connections trained with the SFDFA algorithm align faster and better than those trained with DFA. Moreover, we can observe that the weight alignment correlates with the gradient alignment of the third layer in Figure \ref{fig:sfdfa_gradient_alignment}. This suggests that estimating the forward weights as feedback connections makes the approximate gradients align with the true gradients.

\section{Discussion}

In this paper, we proposed the SFDFA algorithm, a spiking adaptation of FDFA that trains SNNs in an online and local manner. The proposed algorithm computes local gradients of post-synaptic spikes by taking into account all intra-neuron dependencies and uses direct feedback connections to linearly project output errors to hidden neurons without unrolling the neuron's dynamics through space and time. Similarly to the FDFA algorithm, SFDFA estimates the derivatives between hidden and output neurons as feedback connections by propagating directional derivatives as spike grades during the inference. More precisely, SFDFA estimates the weights between output and hidden layers and ignores the temporal relationships between spikes to avoid large variances that could hinder the convergence of feedback connections.
\par
We also demonstrated the existence of critical points where the norm of the exact local gradient diverges towards infinity. These critical points have previously been discovered in other formulations of exact gradients \citep{spikeprop_hair_trigger, spikeprop_surge, event_based_exact_gradient_snn} of spikes. In our work, we identified the cause of these critical points in the computation of the exact local gradients and proposed a simple modification of the derivatives that suppresses gradient explosions. While this ad hoc solution introduces an additional bias to the gradient estimates, we showed that it enabled faster convergence of SNNs than the exact gradient when trained with SFDFA. This implies that the rate of convergence of our algorithm benefits from the improved stability of our modified local gradient.
\par
Our empirical results showed that the proposed SFDFA consistently converges faster and achieves higher performance than DFA on all benchmark datasets. However, while our method also performs better than DFA on the SHD dataset, a significant gap still exists with BP. This suggests that the learning procedure used in both DFA and SFDFA has limitations when applied to highly temporal data. Therefore, future work could explore alternative update methods to improve this performance gap with BP on temporal data.
\par
In our experiments, we measured the alignment between the approximate gradients computed by DFA and SFDFA and the true gradient computed by BP. Our results showed that the approximate gradients computed by the proposed SFDFA algorithm align faster with the true gradient than DFA. This suggests that weights are updated with steeper descending directions in SFDFA than in DFA. This could explain the increased convergence rate experienced by our algorithm. However, we observed that the gradient estimates in SNNs align less than the FDFA algorithm applied to DNNs. This weak alignment can be explained by two factors. First, by bounding the local derivatives, the modified local gradient of our method slightly changes the direction of the weight updates. Second, the complex temporal relationships between spikes are ignored in the computation of the directional derivatives to avoid introducing large variances in the feedback updates. Ignoring these temporal relationships could make the approximate gradients further deviate from the true gradient.
\par
In addition to the gradient alignment, we measured the alignment between the network weights and feedback connections in both DFA and SFDFA. We observed a stronger difference in weight alignment between DFA and SFDFA than in gradient alignment. This could be explained by the fact that our method estimates the weight connections between output and hidden neurons as feedback rather than derivatives. In particular, we observed that the weight alignment of SFDFA correlates with the gradient alignment of the last layer, suggesting that estimating weights contributes to the gradient alignments. However, it is still unclear why deeper layers fail to align more than in DFA. Future work could therefore focus on improving the gradient alignment of deep layers to improve the rate of convergence as well as the performance of SNNs when trained with SFDFA.
\par
From an engineering point of view, the local gradients of spikes can be locally computed by neurons by implementing dedicated circuits that evaluate the dynamical system of the LIF neuron. Moreover, the computation and propagation of directional derivatives during the inference can be implemented through the grades of spikes. Spike grades are features that have recently been added to several large-scale neuromorphic platforms such as Loihi 2 \citep{loihi2} and SpiNNaker \citep{spinnaker}. Traditionally used to modify the amplitude of spikes for computational purposes, our work instead proposes the use of spike grades for learning purposes. Finally, direct feedback connections have widely been implemented on various neuromorphic platforms, thus supporting the hardware compatibility of SFDFA.
\par
Therefore, by successfully addressing the limitations of BP, the proposed SFDFA algorithm represents a promising step towards the implementation of neuromorphic gradient descent. While there are still areas for improvement and exploration, our findings contribute to the growing body of knowledge aimed at improving the field of neuromorphic computing.

\backmatter
\bmhead{Acknowledgements}

This study was part funded by EPSRC grant EP/T008296/1. 
\bmhead{Data and code  availability}

All data used in this paper is publicly available benchmark data and has been cited in the main text \citep{mnist,fashion_mnist,heidelberg_dataset}.

\bmhead{Conflict of interest}

The authors declare that there is no conflict of interest.

\begin{appendices}

\section{Experimental Settings}

In this section, we describe the experimental settings used to produce our empirical results, including the benchmark datasets, encoding and decoding, network architectures, the loss function, hyperparameters as well as the software and hardware settings.

\subsection{Firing Rate Regularization}

Without additional constraint, neurons may exhibit high firing rates to achieve lower loss values which could increase energy consumption and computational requirements. To prevent this issue, we implemented a firing rate regularization that drives the mean firing rate of neurons towards a given target during training. By incorporating firing rate regularization, the neural network is encouraged to find a balance between learning from the data and avoiding high firing rates. This can lead to improved generalization, reduced energy consumption, and enhanced stability.
\par
Formally, a penalty term for high firing rates is added to the loss function, such as:
\begin{equation}
	\mathcal{L}_{\text{reg}}(\boldsymbol{x}) = \mathcal{L}(\boldsymbol{x}) + \beta \sum_{i} \left(\underset{\boldsymbol{x}}{\mathbb{E}}\left[n_i\right] - \widehat{n}\right)^2
\end{equation}
where $\beta > 0$ is a constant defining the strength of the regularization and $\underset{\boldsymbol{x}}{\mathbb{E}}\left[n_i\right]$ is the mean firing rate of the hidden neuron $i$. The mean firing rate can be estimated in an online manner by computing a moving average or an exponential average of the firing rate over the last inference. If batch learning is used, a mean firing rate can be computed from the batch.

\subsection{Update Method and Hyperparameters}

We used the Adam \citep{adam} algorithm to update both the weights and the feedback connections. We used the default values of $\beta_1=0.9$, $\beta_2=0.999$ and $\epsilon=10^{-8}$ \citep{adam} and a batch size of 50 for fully-connected SNNs. We used a learning rate of $\lambda=0.003$ for image classification and $\lambda=0.001$ for audio classification with the SHD dataset. In addition, we used a feedback learning rate of $\alpha=10^{-4}$ in every experiment.
\par
Experimental conditions were standardized for BP, DFA and SFDFA. We used the same hyperparameters as the method proposed in 

\subsection{Event-Based Simulations on GPU}

To simulate and train SNNs, we reused the event-based simulator used in \citep{mypaperwithbacho}.
 We adapted the GPU kernels related to the inference to compute local gradients as well as directional derivatives in an online manner. Moreover, we replaced the code performing error backpropagation with feedback learning. Similarly to neuromorphic hardware, our simulator never backpropagates errors backward through time and performs all computations in an online manner.

\section{Deriving eq.~\ref{eq:loc_grad_final}}

\label{simplified}
 To reduce the computational requirements of the local gradient evaluation, Equation \ref{localgradient} can be further simplified by re-introducing the post-synaptic spike time $t_i^k$ into the equation:
\begin{equation}
	\begin{split}
		\abl{t_i^x }{w_{ij}} 
=& \frac{\exp\left(\frac{-t_i^x}{\tau_s}\right)}{\exp\left(\frac{-t_i^x}{\tau_s}\right)} \frac{\tau}{a_i^x} \left[1 + \frac{\exp\left(\frac{-t_i^x}{\tau}\right)}{\exp\left(\frac{-t_i^x}{\tau}\right)} \frac{\vartheta}{x_i^x} \exp\left(\frac{t_i^x}{\tau}\right)\right] f_i^x - \frac{\exp\left(\frac{-t_i^x}{\tau}\right)}{\exp\left(\frac{-t_i^x}{\tau}\right)} \frac{\tau}{x_i^x} \; h_i^x \\
		=& \frac{\tau}{a_i^x \exp\left(\frac{-t_i^x}{\tau_s}\right)} \left[1 + \frac{\vartheta}{x_i^x \exp\left(\frac{-t_i^x}{\tau}\right)}\right] f_i^x \exp\left(\frac{-t_i^x}{\tau_s}\right) - \frac{\tau}{x_i^x \exp\left(\frac{-t_i^x}{\tau}\right)} \; h_i^x \exp\left(\frac{-t_i^x}{\tau}\right) \\
	\end{split}
\end{equation}
Then we can isolate an expression for $x_i^x \exp\left(\frac{-t_i^x}{\tau}\right)$, such as:
\begin{equation}
	\begin{split}
		&t_i^x = \tau \ln\left(\frac{2a_i^x}{b_i^x + x_i^x}\right) \\
		\Leftrightarrow& \exp\left(\frac{t_i^x}{\tau}\right) = \frac{2a_i^x}{b_i^x + x_i^x} \\
		\Leftrightarrow& \exp\left(\frac{t_i^x}{\tau}\right) \exp\left(\frac{-t_i^x}{\tau}\right) = \exp\left(\frac{-t_i^x}{\tau}\right) \frac{\exp\left(\frac{-t_i^x}{\tau}\right)}{\exp\left(\frac{-t_i^x}{\tau}\right)} \frac{2a_i^x}{b_i^x + x_i^x} \\
		\Leftrightarrow& \exp\left(\frac{t_i^x - t_i^x}{\tau}\right) = \frac{2a_i^x \exp\left(\frac{-t_i^x}{\tau_s}\right)}{\left(b_i^x + x_i^x\right) \exp\left(\frac{-t_i^x}{\tau}\right)} \\
		\Leftrightarrow& 1 = \frac{2a_i^x \exp\left(\frac{-t_i^x}{\tau_s}\right)}{\left(b_i^x + x_i^x\right) \exp\left(\frac{-t_i^x}{\tau}\right)} \\
		\Leftrightarrow& x_i^x \exp\left(\frac{-t_i^x}{\tau}\right) = a_i^x \exp\left(\frac{-t_i^x}{\tau_s}\right) - \vartheta
	\end{split}
\end{equation}
as of the constraint between the synaptic and membrane time constant $\tau = 2\tau_s \Leftrightarrow \exp\left(\frac{-t_i^x}{\tau}\right)^2 = \exp\left(\frac{-t_i^k}{\tau_s}\right)$.
\begin{equation}
	\begin{split}
		&u(t_i^x) = b_i^x \exp\left(\frac{-t_i^x}{\tau}\right) - a_i^x \exp\left(\frac{-t_i^x}{\tau_s}\right) = \vartheta \\ 
		\Leftrightarrow& b_i^x \exp\left(\frac{-t_i^x}{\tau}\right) = a_i^x \exp\left(\frac{-t_i^x}{\tau_s}\right) + \vartheta
	\end{split}
\end{equation}
Therefore, the local gradient $\abl{t_i^x }{w_{ij}}$ of spikes simplifies to:
\begin{equation} 
	\begin{split}
		\abl{t_i^x }{w_{ij}} =& \frac{\tau}{a_i^x \exp\left(\frac{-t_i^x}{\tau_s}\right)} \left[\frac{a_i^x \exp\left(\frac{-t_i^x}{\tau_s}\right) - \vartheta}{a_i^x \exp\left(\frac{-t_i^x}{\tau_s}\right) - \vartheta} + \frac{\vartheta}{a_i^x \exp\left(\frac{-t_i^x}{\tau_s}\right) - \vartheta}\right] f_i^x \exp\left(\frac{-t_i^x}{\tau_s}\right) \\ 
		& - \frac{\tau}{a_i^x \exp\left(\frac{-t_i^x}{\tau_s}\right) - \vartheta} h_i^x \exp\left(\frac{-t_i^x}{\tau}\right) \\
		=& \frac{\tau}{a_i^x \exp\left(\frac{-t_i^x}{\tau_s}\right) - \vartheta} \left[f_i^x \exp\left(\frac{-t_i^x}{\tau_s}\right) - h_i^x \exp\left(\frac{-t_i^x}{\tau}\right)\right]
	\end{split}   
\end{equation}

\end{appendices}

 

\end{document}